%% file: WWW2026/main/main.tex
\documentclass[sigconf]{acmart}
\makeatletter
\@ifundefined{Bbbk}{}{%
  
}
\makeatother

\AtBeginDocument{%
  }

\copyrightyear{2026}
\acmYear{2026}
\setcopyright{cc}
\setcctype{by}
\acmConference[WWW '26]{Proceedings of the ACM Web Conference 2026}{April 13--17, 2026}{Dubai, United Arab Emirates}
\acmBooktitle{Proceedings of the ACM Web Conference 2026 (WWW '26), April 13--17, 2026, Dubai, United Arab Emirates}
\acmPrice{}
\acmDOI{10.1145/3774904.3793046}
\acmISBN{979-8-4007-2307-0/2026/04}
  
\acmISBN{978-1-4503-XXXX-X/2018/06}

\usepackage{siunitx}
\usepackage{algorithm}
\usepackage{algorithmic}
\usepackage{appendix}
\usepackage{booktabs}
\usepackage{amsmath}
\usepackage{amssymb}
\usepackage{multicol}
\usepackage{multirow}
\usepackage{array}
\usepackage{bm}
\usepackage{pbox}
\usepackage{balance}
\usepackage{afterpage}
\usepackage{float} 
\usepackage{blindtext}
\usepackage{graphicx}
\graphicspath{{./graphics/}}
\DeclareGraphicsExtensions{.pdf,.png,.jpg}
\usepackage{multicol} % for multi-column layout
\usepackage{makecell}
\newcolumntype{P}[1]{>{\centering\arraybackslash}p{#1}}
\usepackage{xcolor}

\usepackage[belowskip=2pt,aboveskip=0pt]{caption}
\usepackage{hyperref}

\begin{document}

\title{StreamSense: Streaming Social Task Detection with Selective Vision–Language Model Routing}
%\title{StreamSense: A Bi-Model Encoder–VLM Framework for Efficient Streaming Social Task Detection}


% \author{Han Wang}
% \orcid{0009-0007-4486-0693}
% \affiliation{
%  \institution{Singapore University of Technology and Design}}
%   % \country{Singapore}
% \email{han_wang@sutd.edu.sg}

% \author{Deyi Ji}
% \orcid{0000-0001-7561-9789}
% \affiliation{
%  \institution{University of Science and \\ Technology of China}}
% \email{jideyi@mail.ustc.edu.cn}

% \author{Lanyun Zhu}
% \orcid{0000-0001-7309-3330}
% \affiliation{%
%   \institution{Nanyang Technological Universities}}
% \email{lanyun_zhu@ntu.edu.sg}

% \author{Jiebo Luo}
% \orcid{0000-0002-4516-9729}
% \affiliation{%
%   \institution{University of Rochester}}
% \email{jiebo.luo@gmail.com}

% \author{Roy Ka-Wei Lee}
% \orcid{0000-0002-1986-7750}
% \affiliation{%
%   \institution{\makebox[0pt][c]{\mbox{Singapore University of Technology and Design}}}
% }
% \email{s.roylee@gmail.com}
\author{Han Wang}
\orcid{0009-0007-4486-0693}
\email{han_wang@sutd.edu.sg}
\affiliation{%
  \institution{Singapore University of Technology and Design}
  \country{Singapore}
}

\author{Deyi Ji}
\orcid{0000-0001-7561-9789}
\email{jideyi@mail.ustc.edu.cn}
\affiliation{%
  \institution{University of Science and Technology of China}
  \country{China}
}

\author{Lanyun Zhu}
\orcid{0000-0001-7309-3330}
\email{lanyun_zhu@ntu.edu.sg}
\affiliation{%
  \institution{Nanyang Technological University}
  \country{Singapore}
}

\author{Jiebo Luo}
\orcid{0000-0002-4516-9729}
\email{jiebo.luo@gmail.com}
\affiliation{%
  \institution{University of Rochester}
  \country{USA}
}

\author{Roy Ka-Wei Lee}
\orcid{0000-0002-1986-7750}
\email{s.roylee@gmail.com}
\affiliation{%
  \institution{Singapore University of Technology and Design}
  \country{Singapore}
}
\renewcommand{\shortauthors}{Han Wang, Deyi Ji, Lanyun Zhu, Jiebo Luo, and Roy Ka-Wei Lee}
%% No italics, no superscripts, not anonymous
%% Use footnote or author note to identify equal contribution and/or contact author info


\begin{abstract}
Live streaming platforms require real-time monitoring and reaction to social signals, utilizing partial and asynchronous evidence from video, text, and audio. We propose \textsf{StreamSense}, a streaming detector that couples a lightweight streaming encoder with selective routing to a Vision–Language Model (VLM) expert. \textsf{StreamSense} handles most timestamps with the lightweight streaming encoder, escalates hard/ambiguous cases to the VLM, and defers decisions when context is insufficient. The encoder is trained using (i) a cross-modal contrastive term to align visual/audio cues with textual signals, and (ii) an IoU-weighted loss that down-weights poorly overlapping target segments, mitigating label interference across segment boundaries. We evaluate \textsf{StreamSense} on multiple social streaming detection tasks (e.g., sentiment classification and hate content moderation), and the results show that \textsf{StreamSense} achieves higher accuracy than VLM-only streaming while only occasionally invoking the VLM, thereby reducing average latency and compute. Our results indicate that selective escalation and deferral are effective primitives for understanding streaming social tasks. Code is publicly available on \href{https://github.com/Social-AI-Studio/StreamSense}{GitHub}.

%With the rapid growth of social media, live-streaming videos have become a major medium for real-time human interaction, driving the need for \textit{streaming social task detection}. Unlike conventional video understanding, streaming social content is multimodal and requires real-time interpretation of complex human behaviors using only historical context. 
%We investigate real-time detection on social streaming videos and identify two key challenges: (1) balancing performance and efficiency and (2) dealing with the absence of streaming annotations, which leads to label interference across video segments and the need to make predictions with limited context.
%To address these issues, we propose %StreamSense, 
%StreamSense, 
%a bi-model framework that integrates a lightweight multimodal transformer with a large vision–language model (VLM). The Encoder efficiently handles most easy predictions with a cross-modal contrastive loss to align modalities, while an IoU-based cross-entropy loss accounts for partial segment visibility to mitigate label interference.
%For difficult predictions with limited context, we propose a prediction defer strategy, where the VLM decides when to defer due to limited context and takes over.
%Experiments show our model outperforms fully VLM-based methods with only 27\% VLM usage (17\% for defer decisions) and a latency of \SI{0.3}{\second} per prediction.
\end{abstract}



% <concept_id>10010147.10010178.10010179</concept_id>
%        <concept_desc>Computing methodologies~Natural language processing</concept_desc>
%        <concept_significance>500</concept_significance>
%        </concept>
%    <concept>
% \ccsdesc[500]{Computing methodologies~Natural language processing}




\maketitle

\section{Introduction}

\input{main/introduction}

\section{Related Works}
\input{main/related}

\section{Streaming Social Task Detection}
\input{main/task}

\section{Methodology}

\input{main/model}

\section{Experiments}

\input{main/experiments}

% \section{Case Study}
% \input{main/case}

\section{Conclusion}
\input{main/conclusion}

\section*{Acknowledgement}
This research is supported in part by the National Research Foundation, Prime Minister’s Office, Singapore, and the Ministry of Digital Development and Information, under its Online Trust and Safety (OTS) Research Programme (Award Grant No. S24T2TS007). Any opinions, findings, and conclusions or recommendations expressed in this material are those of the author(s) and do not reflect the views of the National Research Foundation, Prime Minister’s Office and  Singapore, or the Ministry of Digital Development and Information, Singapore.

\bibliographystyle{ACM-Reference-Format}
\bibliography{main/ref}

\end{document}

%% file: main/introduction.tex
\begin{figure}[t]
  \centering
  \begin{minipage}{0.48\textwidth}
    \centering
    \includegraphics[width=\linewidth]{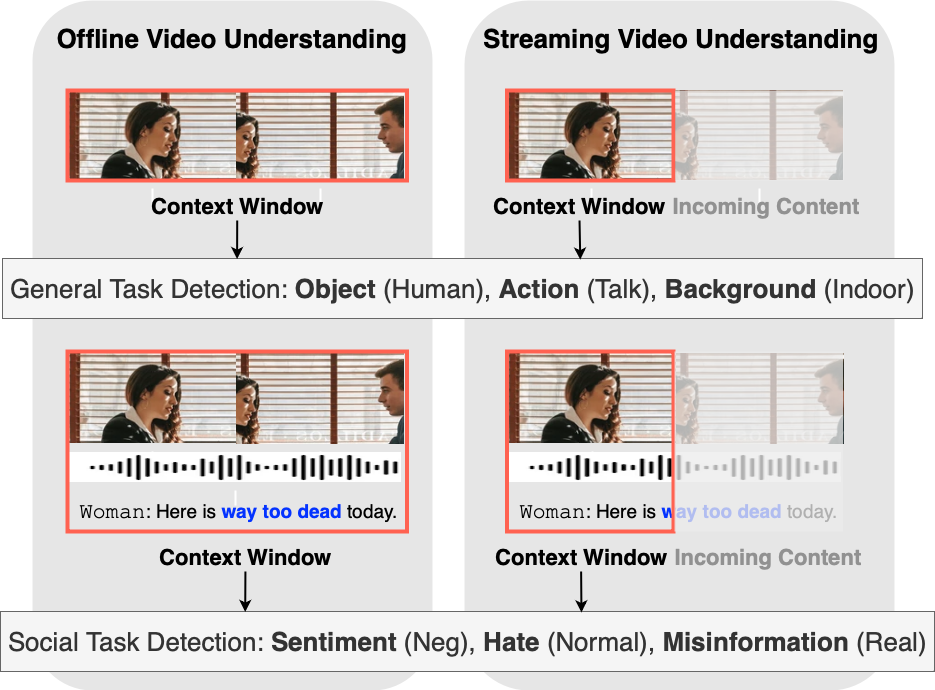}
  \end{minipage}
    \caption{Comparison of (1) offline versus streaming video understanding and (2) general versus social task detection. Neg = Negative, Hate = Hate Speech.}
     \label{fig:task_comparison}
\end{figure}

Live streaming platforms (e.g., Twitch, YouTube Live, TikTok Live, etc.) deliver low-latency video to millions while enabling real-time interaction, creating a tight feedback loop between content creators and their audiences. These platforms are popular for gaming, entertainment, live shopping, education, breaking news, and civic events, and underpin the creator economy. In this setting, understanding social signals as they emerge is crucial: streams exhibit rapid sentiment shifts, harassment waves, hate raids, coordinated misinformation during evolving events, deceptive promotions, and context-dependent toxicity. These signals are multimodal (spoken/on-screen text often provides decisive cues while visual and audio signals provide context) and asynchronous, complicating instant interpretation. Effective moderation, safety interventions, and analytics, therefore, require real-time inference under partial context; deciding when current evidence suffices, when to wait for more, and when to escalate to stronger models.

Therefore, we formalize \textit{streaming social task detection} (SSTD) as the problem of making a decision at each timestamp using past-only evidence from video, text, and audio. As illustrated in Figure~\ref{fig:task_comparison}, SSTD differs from conventional (offline/general) video detection along three axes: 
(i) \textbf{Causality \& latency}: predictions cannot peek into future content and must be synchronized with the live stream; (ii) \textbf{Multimodality}: while general task detection is often vision-centric, social tasks frequently hinge on spoken or on-screen language with visuals providing situational context; and 
%\textcolor{red}{(iii) \textbf{Social inference}: labels such as intent, stance, or toxicity may require non-local reasoning and background knowledge rather than a visually localized trigger.}  
(iii) \textbf{Social inference}: detection requires substantial human understanding and background knowledge to interpret intent, stance, or toxicity beyond visually localized cues.
These differences make offline or general task methods unfit for streaming social analysis.

Prior work on offline social tasks typically couples frozen encoders with deep classifiers~\cite{zadeh-etal-2017-tensor,liang2018multimodal,wang2019words,hazarika2020misa,yu2021learning,wang2023tetfn}. With the rise of Large Language Models (LLMs) and Vision-Language Models (VLMs), recent studies show strong performance on social nuanced tasks due to broad pretraining on social media and web content~\cite{zhang2023sentiment,guo2023investigation,li2024agent,nirmal2024interpretable,pendyala2024explaining}. However, directly applying LLM/VLM systems to SSTD faces two obstacles: (1) \textbf{integration}: bringing new modality streams into large models typically requires costly multimodal pretraining, and (2) \textbf{efficiency}: high-capacity models are computationally expensive and slow for continuous, large-scale, real-time inference.

Streaming work in general task detection is mainly focused on online action recognition (OAR)~\cite{degeest2016online,wang2021oadtr,xu2021long,gao2021woad,xu2019temporal}. Those labels are time-synchronous: When the action occurs, the ground truth is immediately knowable. In contrast, for social tasks, annotators often cannot commit at a given timestamp without observing full context. Proper stream-level supervision should therefore allow an \textsf{Unknown} state until sufficient evidence accrues (Figure~\ref{fig:Stream_Segment_Annotation}). In practice, available datasets only provide segment-level labels created after viewing the whole segment, which induces two pathologies for streaming models: (a) label interference between adjacent windows with conflicting segment labels in training-time, and (b) forced predictions under insufficient context. Unsurprisingly, naively adapting OAR architectures to SSTD yields weak results~\cite{wang2025hateclipseg}.

To address these research gaps, we propose \textsf{StreamSense}, a real-time multimodal system that combines a lightweight streaming encoder with selective routing to a high-capacity VLM expert, and a learned deferral policy when context is insufficient. The streaming encoder ingests past-only video, text and audio; a router estimates probability score and either (i) emits a prediction, (ii) escalates the current window to the VLM, or (iii) defers until more evidence arrives. To train robustly under segment-level supervision, we use (i) a cross-modal contrastive objective to align visual/audio cues with decisive textual signals, and (ii) an IoU-weighted cross-entropy that down-weights windows with low overlap to the target segment, mitigating label interference across adjacent windows. This design outperforms VLM-only baselines while sparingly invoking the VLM, yielding lower latency and compute.

We summarize our contributions as follows: (1) We formalize \textit{streaming social task detection} and articulate why it fundamentally differs from conventional video detection in terms of causality, multimodality, and social inference; (2) We introduce \textsf{StreamSense}, a selective-routing framework that combines a lightweight streaming encoder with an on-demand VLM expert and a deferral policy for low-context timestamps; and (3) We address supervision mismatch with IoU-weighted classification to reduce label interference and a cross-modal contrastive alignment to enhance modality fusion, achieving strong accuracy while substantially reducing VLM calls.

%% file: main/related.tex
\subsection{Social Task Detection}

Social task detection studies human behaviors and intentions in social media content across modalities (text, images/memes, and video). We focus on video settings with representative subtasks, including sentiment/affect classification, hate/harassment detection, and misinformation identification.

Sentiment is the most resourced area, with IEMOCAP~\cite{busso2008iemocap}, CMU-MOSI~\cite{zadeh2016mosi}, CMU-MOSEI~\cite{zadeh2018cmumosei}, MELD~\cite{poria2018meld}, CH-SIMS~\cite{yu2020ch}, and CMU-MOSEAS~\cite{zadeh2020cmumoseas} offering video- or segment-level annotations (sometimes fine-grained emotions). MOSI/MOSEI are particularly popular due to their strict guidelines, which ensure each segment conveys a complete and consistent opinion. Hate/harassment datasets such as HateMM~\cite{das2023hatemm}, MultiHateClip~\cite{wang2024multihateclip}, HateClipSeg~\cite{wang2025hateclipseg}, and ImpliHateVid~\cite{rehman2025implihatevid} provide video/segment labels and, in some cases, victim categories. 
%\textcolor{red}{However, outside HateClipSeg, segment boundaries are often subjectively defined, complicating agreement and temporal grounding.} 
However, outside HateClipSeg, segment boundaries are subjective and may fail to capture complete or consistent opinions. Misinformation video resources remain scarce~\cite{papadopoulou2019corpus,boididou2018misleading} and typically lack frame/segment alignment. Finally, these segment-level corpora are suited for offline analysis but not for stream-level supervision (i.e., lacking an \textsf{Unknown} state).

Traditional social video detection and understanding models fuse unimodal features using deep architectures~\cite{zadeh-etal-2017-tensor,liang2018multimodal,wang2019words,dlpl,hazarika2020misa,yu2021learning,wang2023tetfn,wang_video_image_sentence_YYYY}. Canonical examples include TFN~\cite{zadeh-etal-2017-tensor} (tensor fusion of uni/bi/trimodal interactions), MISA~\cite{hazarika2020misa} (modality-invariant/specific factors), and Self-MM~\cite{yu2021learning} (self-supervised multimodal pretraining). Recent work leverages LLMs for sentiment~\cite{zhang2023sentiment}, hate~\cite{guo2023investigation}, and misinformation~\cite{li2024agent}, with advantages in zero-/few-shot generalization and explanation~\cite{wang2025multi,wang2023gpt3,raven,raven++,nirmal2024interpretable,zhucpcf,pendyala2024explaining,cao_procap_2023,lu_llm_offensive_2025}. Yet, integrating visual/audio into language models typically requires heavy multimodal pretraining, 
%\textcolor{red}{and per-timestamp inference is computationally expensive for continuous, high-rate monitoring.} 
and the high inference cost of language models hinders per-timestamp analysis in continuous, high-frequency monitoring. 

%\textcolor{red}{Most of the aforementioned social detection models are trained on offline and trimmed offline segments with full context and (often) synchronized modalities. In livestreams, evidence arrives asynchronously, and causality forbids peeking, so sliding-window use yields premature or unstable labels. Segment-level supervision also forces an always-decide behavior, causing boundary interference where the correct state at time $t$ is effectively Unknown. \textsf{StreamSense} addresses this gap with a lightweight streaming encoder, selective VLM escalation, and an explicit deferral policy, trained to make predictions with partial context.}
Most of the aforementioned social detection models do not require real-time inference and are trained on offline, trimmed segments with full context. In livestreams, real-time detection requires a high inference rate, making full reliance on LLMs/VLMs prohibitively costly. Moreover, livestreams require causal processing, so evidence may arrive asynchronously; forcing immediate decisions can yield premature labels under insufficient context. \textsf{StreamSense} addresses this gap with a lightweight streaming encoder, selective VLM escalation, and an explicit deferral policy for limited context.%} \textcolor{orange}{Label interference is mitigated via the IoU-based loss; if the solution is not described here, the problem should likewise be omitted here.}

\subsection{Streaming Video Understanding}
Streaming video understanding typically targets online action recognition (OAR), where models emit human action labels in real time~\cite{degeest2016online,wang2021oadtr,xu2021long,gao2021woad,xu2019temporal}. The emphasis is on temporal modeling under causality: OadTR~\cite{wang2021oadtr} captures global dependencies with Transformers; LSTR~\cite{xu2021long} blends long/short-term cues; WOAD~\cite{gao2021woad} learns from weak labels via pseudo frame supervision. However, both OAR and recent streaming highlight prediction tasks~\cite{deng2024multimodal} assume temporally aligned supervision. %\textcolor{red}{In contrast, social tasks are asynchronous: labels (e.g., sentiment, toxicity) may be unknowable at a timestamp until subsequent evidence appear, which leads to label interference when training from post-hoc segment labels.}
In contrast, social tasks are often asynchronous: labels (e.g., sentiment, hate, misinformation) may be unknowable at a given timestamp but are typically assigned using future content in segment-level annotations, which can create consecutive windows with similar features yet conflicting labels, leading to label interference during training.

A complementary line of works explores streaming visual question answering (VQA) ~\cite{chen2024videollm,zhang2024flashvstream,di2025streaming,wang2025streambridge}. To improve efficiency, works explore parallel encoders and asynchronous LLM inference~\cite{chen2024videollm}, memory mechanisms for long contexts~\cite{zhang2024flashvstream}, and cache designs for streaming~\cite{di2025streaming}. However, VQA is query-driven and low-rate; it does not mandate \emph{high-frequency} decisions at every timestamp. Directly applying such models to social monitoring yields limited gains relative to their computational and fine-tuning costs.

\textsf{StreamSense} differs from online action recognition and streaming VQA. We target high-cadence social monitoring with asynchronous evidence across video, text and audio. Our design (i) uses a lightweight streaming encoder for the typical case, (ii) selectively routes hard cases to a fine-tuned offline VLM expert, and (iii) defers when context is insufficient. Encoder training uses IoU-weighted classification and cross-modal contrastive losses to reduce label interference and stabilize multimodal fusion under partial context.

%% file: main/task.tex
\begin{figure}[t]
  \centering

  %\begin{minipage}{0.58\textwidth}
    \centering
    \includegraphics[width=\linewidth]{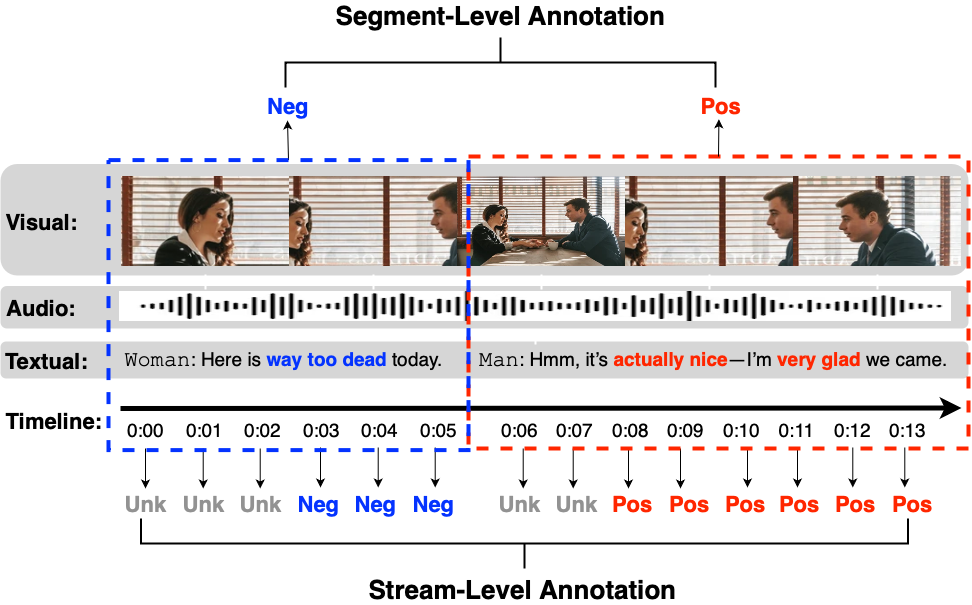}
  %\end{minipage}
    \caption{Comparison of segment-level and stream-level annotations using sentiment classification as an example. Pos = Positive, Neg = Negative, Unk = \textsf{Unknown}}
     \label{fig:Stream_Segment_Annotation}
\end{figure}

In this section, we first formalize \textit{streaming social task detection} by specifying the inputs, outputs, and evaluation criteria. We then highlight two core challenges that arise in livestream settings: (i) achieving high accuracy under tight latency and compute budgets, and (ii) coping with supervision that is segment-level and post hoc rather than truly stream-level.
% —and preview the design choices we adopt to address them.

\subsection{Task Formulation}
We define \textit{streaming social task detection} (SSTD) as real-time inference of social labels (e.g., sentiment, hate speech, misinformation) from a live, multimodal video stream. Let the stream be features $[\mathbf{x}_1,\ldots,\mathbf{x}_T]$ over $T$ time steps, where each $\mathbf{x}_i=(v_i,t_i,a_i)$ comprises visual, textual, and audio cues available up to time $i$.

\noindent\textbf{Streaming prediction (past-only).}
%At fixed decision intervals $s$ (e.g., every frame or every $\Delta$ seconds), the model predicts a label using a past-only window of length $N$:
%\[
%\hat{y}_i = f(\mathbf{x}^i), \qquad \mathbf{x}^i=[\mathbf{x}_{i-N},\ldots,\mathbf{x}_i], \qquad i \in \{s,2s,\ldots,T\},
%\]
At fixed decision intervals $s$ (e.g., every $\Delta$ seconds), the model predicts a label using past-only window of length $N$:
\[ f(\mathbf{x}^i) \to \hat{y}_i, \qquad \mathbf{x}^i=[\mathbf{x}_{i-N},\ldots,\mathbf{x}_i], \qquad i \in \{s,2s,\ldots,T\}.
\]
with frames $j<i-N$ outside the window and frames $j>i$ unseen (for $i<N$, we use fixed padding). Unlike offline classification, SSTD forbids peeking beyond time $i$, requires per-interval decisions from a bounded past-only window $[\mathbf{x}_{i-N},\ldots,\mathbf{x}_i]$, and must fuse video-text–audio signals that can be temporally misaligned (e.g., speech trailing the visuals), so some labels may be undecidable at time $i$ until later evidence appears.

\noindent\textbf{Supervision \& Evaluation.}
We next describe how existing datasets supervise SSTD and how we evaluate it. Many social video corpora provide segment-level labels: a segment $[st,en]$ is annotated post hoc as a consistent opinion/state $y_{st:en}\!\in\!\mathcal{Y}$. In the trimmed setting, a classifier consumes the whole segment,
\[
f(\mathbf{x}_{st:en}) \to \hat{y}_{st:en}, \qquad \mathbf{x}_{st:en}=[\mathbf{x}_{st},\ldots,\mathbf{x}_{en}].
\]

For SSTD, frame/timestamp labels are derived by propagation, $y_i = y_{st:en}$ for $i \in [st,en]$, and we measure accuracy and macro-F1 (M-F1) of $\hat{y}_i$ using a past-only window $[\mathbf{x}_{i-N},\ldots,\mathbf{x}_i]$. %\textcolor{red}{This post-hoc supervision lacks an \textsf{Unknown} state to indicate ``insufficient evidence at time $i$'', creating a supervision–inference mismatch addressed next.}
This post-hoc supervision lacks an \textsf{Unknown} state to indicate ``insufficient evidence at time $i$''  and serve as ``a buffer between consecutive windows with conflicting labels'', creating a supervision–inference mismatch addressed next.

\subsection{Challenge I: Balancing Performance and Efficiency} SSTD requires frequent, low-latency decisions while fusing video, text, and audio using only past evidence. At stream cadence, inputs arrive as micro-batches, limiting the ability to batch efficiently and exponentially increasing the per-timestamp cost of inference for high-capacity models. Beyond inference, multimodal pre-processing is nontrivial: automatic speech recognition and on-screen text extraction introduce latency and jitter relative to video frames, and keeping these streams synchronized online adds computational and memory overhead, consuming the same real-time budget. Evidence for social labels is also sparse and asynchronous. Decisive cues (a short phrase, a brief overlay, or a fleeting visual) often appear after an initially ambiguous moment in the scene. Running heavy inference uniformly at every timestamp wastes computation on windows where the label is genuinely undecidable, yet naive throttles (larger stride, frame dropping) systematically miss short events and reduce recall. Standard confidence heuristics are unreliable in partial or misaligned contexts: deep classifiers can be overconfident on non-informative inputs, and calibration tuned offline does not transfer cleanly to stream-time uncertainty.

Simple remedies such as fixed downsampling, static ASR/OCR cadence, early-exit rules borrowed from single-modality settings, or one-shot distillation into a small model, do not address the underlying constraints. They either degrade tail performance on rare but safety-critical moments, cause performance drift across topics and languages, or remain operationally brittle. In summary, SSTD sits at a compute–quality frontier defined by high decision frequency, costly multimodal pre-processing, asynchronous and sparse evidence, and strict latency/compute budgets. These factors render the task more challenging than trimmed or offline detection and are not amenable to simple modifications of existing pipelines.

\subsection{Challenge II: Absence of Stream-Level Annotation}
SSTD is evaluated at stream time; however, available corpora provide only post-hoc segment-level supervision (Fig.~\ref{fig:Stream_Segment_Annotation}). Propagating a segment label $y_{st:en}$ to every timestamp $i\in[st,en]$ implicitly assumes that decisive evidence is uniformly distributed across the segment. In practice, social cues are sparse and asynchronous: the utterance or text that determines the label may occur late, while earlier windows remain undecidable. This supervision–inference mismatch produces several technical challenges. 
First, near segment boundaries, consecutive windows may appear identical from past-only features yet receive different labels; training a causal model to be confident in both creates conflicting gradients and unstable decision surfaces, especially for lightweight encoders. 
Second, the propagated labels encourage an ``always decide'' behavior at timestamps, even in inherently undecidable cases. Standard cross-entropy offers no mechanism for representing those \textsf{Unknown} cases, and calibration learned offline does not transfer reliably to partial, misaligned context. 
Third, propagation amplifies annotator subjectivity and temporal drift, as minor boundary shifts introduce systematic noise that cannot be smoothed without look-ahead, violating real-time constraints. This necessitates strict annotation policies and rigorous quality control.
Fourth, multimodal asynchrony exacerbates all of the above: speech and text tokens often lag the visual stream, so windows aligned by frame index are misaligned in evidential content, yet are trained as if fully observed.

Common ``lightweight fixes'' do not resolve these issues. Expanding the context window or dilating labels reduces boundary churn but increases latency and memory, and still forces premature decisions when evidence is absent rather than merely weak. Label smoothing or tolerance bands reduce overconfidence, but they also blur class boundaries and degrade the detection of brief, safety-critical events. Reweighting losses near boundaries is hard to do without privileged knowledge of where evidence actually occurs; uniform heuristics either under-correct or introduce bias toward longer segments. Pseudo-labeling or self-training with the same propagated supervision tends to reinforce the original errors. As a result, the lack of true stream-level labels with an explicit \textsf{Unknown} state is not a minor nuisance but a structural obstacle: it couples training instability near boundaries with inference-time pressure to guess, precisely where causal evidence is insufficient.

%% file: main/model.tex
\begin{figure*}[t]
  \centering
  \begin{minipage}{0.9\textwidth}
    \centering
    \includegraphics[width=\linewidth]{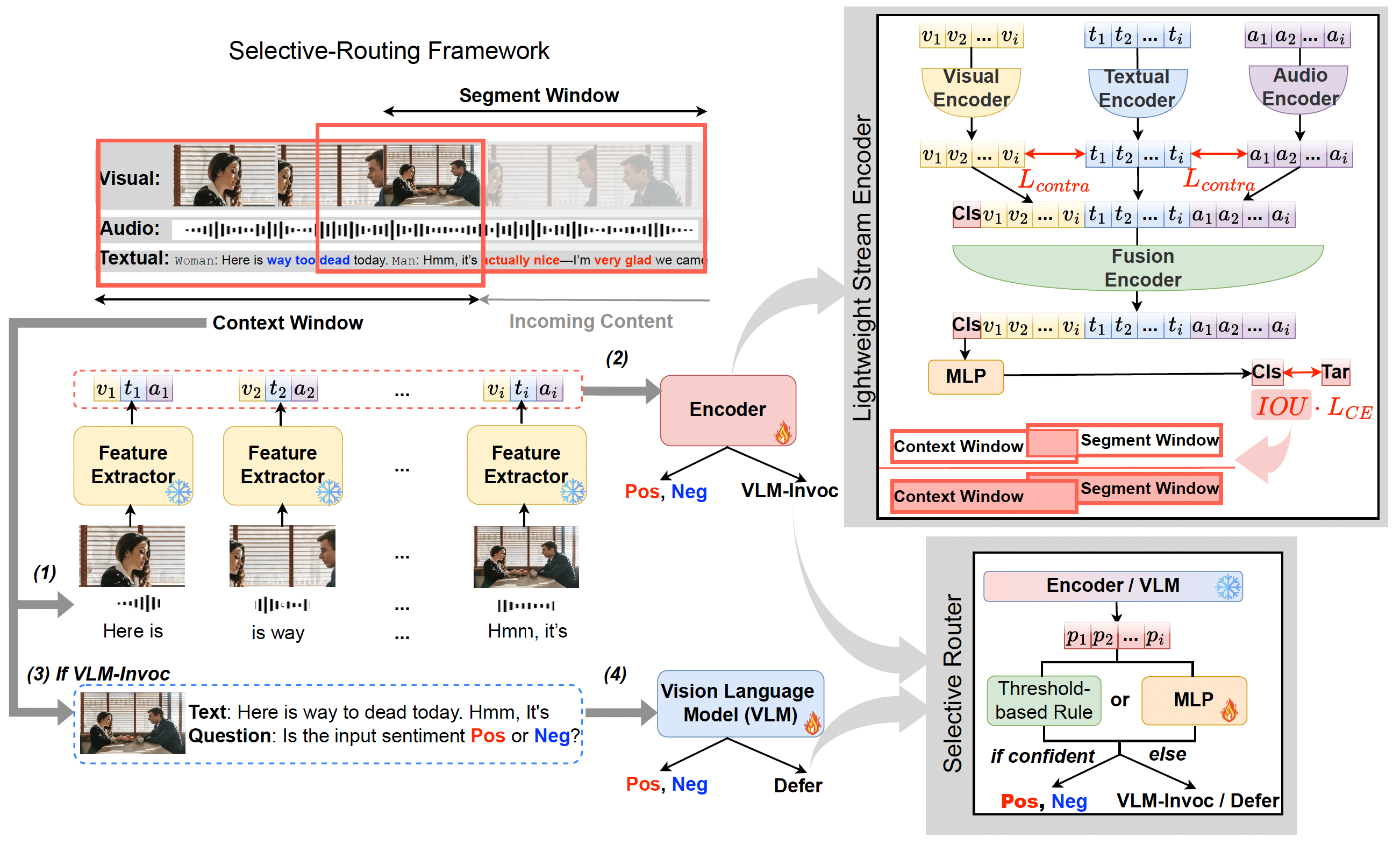}
  \end{minipage}
    
    \caption{Overview of the proposed \textsf{StreamSense} framework.}
    \label{fig:overall_framework}
\end{figure*}

This section presents \textsf{StreamSense} and its core components. This streaming detector integrates a lightweight Streaming Encoder with \emph{selective routing} to a high-capacity VLM expert and an explicit deferral policy. Figure~\ref{fig:overall_framework} provides an overview of the \textsf{StreamSense} framework. Given a live stream, unimodal feature extractors produce per-timestamp features $\mathbf{x}_i = (v_i,t_i,a_i)$. The Streaming Encoder consumes a past-only window $[\mathbf{x}_{i-N},\ldots,\mathbf{x}_i]$ and outputs a provisional label $\hat{y}_i$ and predicted class probability $p_i$. A \textit{Router} then chooses among three actions at time $i$: \textit{emit} the encoder’s label, \textit{escalate }to the VLM expert (which re-analyzes inputs up to $i$), or \textit{defer} the decision when evidence is insufficient. The encoder is trained with (i) a cross-modal contrastive objective to stabilize alignment across modalities and (ii) an IoU-weighted classification loss to reduce boundary interference under segment-level supervision. These core components are discussed in the following sections.

%This section presents \textsf{StreamSense}, a streaming detector that integrates a lightweight encoder with selective VLM routing for \textit{streaming social task detection}. The overall framework is shown in Figure~\ref{fig:overall_framework}, with details of the selective-routing framework, lightweight stream encoder classifier, and VLM-invocation and prediction defer strategy presented below. For a streaming video input, unimodal feature extractors tokenize each modality at timestamp $i$ into $x_i = [v_i, t_i, a_i]$. A lightweight stream encoder classifier processes features within the context window to provide an initial prediction, trained using cross-modal contrastive loss and IoU–based cross-entropy loss. Under the VLM-invocation strategy, in-confident Encoder predictions with complex cases trigger a finetuned VLM to reprocess the raw input within the context window for the final prediction, while a prediction defer strategy is employed when the VLM is not confident, given the limited context.

\subsection{Streaming Encoder (Classifier)}

As shown in Figure~\ref{fig:overall_framework}, the lightweight stream encoder consists of three unimodal encoder branches for visual, textual, and audio inputs. For timestamp $i$, the unimodal sequences
$v^i = [v_{i-N}, \ldots, v_i]$, $t^i = [t_{i-N}, \ldots, t_i]$, and $a^i = [a_{i-N}, \ldots, a_i]$
are first processed by an $l_{enc}$-layer encoder for each modality, followed by an $l_{fus\_enc}$-layer fusion encoder. The resulting unimodal representations are concatenated and fed into a multi-layer perceptron (MLP) for prediction. Training follows the streaming setting with two objectives: (i) a cross-modal contrastive loss for modality alignment and (ii) an IoU-based cross-entropy loss to mitigate label interference across consecutive segments.

\noindent\textbf{Cross-modal contrastive alignment.}
The cross-modal contrastive loss aligns visual ($v^i$) and audio ($a^i$) features with textual features ($t^i$), as text commonly provides the most salient information in social-task settings:
\begin{equation}
\mathcal{L}^\alpha_{\text{CM}} = \frac{\alpha}{2} \Big( \mathcal{L}_{\text{contra}}(t, v) + \mathcal{L}_{\text{contra}}(t, a)\Big),
\end{equation}
where $\alpha$ controls the weight of the cross-modal term ($\alpha{=}0$ indicates no contribution), and

\begin{equation}
\mathcal{L}_{\text{contra}}(x, y) = - \frac{1}{B} \sum_{i=1}^{B} \log \frac{\exp(s_{i,i} / \tau)}{\sum_{j \neq i} \exp(s_{i,j}/ \tau) + \exp(s_{i,i} / \tau)}
\end{equation}
where $ \quad s_{i,j} = \text{sim}(x^i, y^j)$, with $\mathrm{sim}(\cdot,\cdot)$ denoting cosine similarity, $B$ the batch size, and $\tau$ a temperature.
Intuitively, decisive cues in social tasks are often verbal or on-screen text, while visual and acoustic signals provide context or precursors (facial affect, prosody, scene). Because modalities arrive at different rates and can be temporarily misaligned, the encoder may over-rely on whichever stream is easiest at a given moment and underuse others. 
The contrastive term aligns visual and audio representations with text at the same window, distancing them from representations from different timestamps or videos.
This alignment makes cross-modal fusion more stable when one modality is weak or missing, and reduces sensitivity to transient misalignment. 
The scalar $\alpha$ controls how strongly this coupling is enforced relative to the supervised objective.

\noindent\textbf{IoU-based cross-entropy.}
We define an IoU-weighted cross-entropy loss that adaptively scales the contribution of each temporal instance according to the temporal overlap between the context window and the ground-truth segment:
\begin{equation}
\mathcal{L}_{\text{IoU * CE}}^\beta
= - \frac{1}{B} \sum_{i=1}^{B}
\mathrm{IoU}(W^i, S^i)^\beta \cdot y_i \cdot \log p(y_i \mid x^i),
\end{equation}
where
\begin{equation}
\mathrm{IoU}(W^i, S^i)
= \frac{|W^i \cap S^i|}{|W^i \cup S^i|}.
\end{equation}
Here, \( W^i = [W_{st}, W_{en}] \) and \( S^i = [S_{st}, S_{en}] \) denote temporal boundaries, and \( \beta \) controls the strength of IoU weighting, with \( \beta = 0 \) reducing to standard cross-entropy.

% The IoU-based cross-entropy loss modulates the contribution of each timestamp according to the temporal intersection-over-union (IoU) between the current context window and the labeled segment at time $i$:
% % \end{equation}
% \begin{equation}
% \mathcal{L}_{\text{IoU * CE}}^\beta
% = - \frac{1}{B} \sum_{i=1}^{B}
% \mathrm{IoU}(W^i, S^i)^\beta \cdot y_i \cdot \log p(y_i \mid x^i)
% \end{equation}
% \begin{equation}
% \mathrm{IoU}(W^i, S^i)
% = \frac{|W^i \cap S^i|}{|W^i \cup S^i|}
% \end{equation}
% where $W^i = [W_{st}, W_{en}]$ and $S^i = [S_{st}, S_{en}]$ represent the current context and segment window boundaries, $p(y_i \mid x^i)$ is the predicted probability, and $\beta$ controls the influence of the IoU weight on cross-entropy, with $\beta=0$ corresponding to standard cross-entropy.

The motivation is that segment labels are assigned after seeing the full segment, whereas streaming decisions are made before all evidence is observed. Propagating a segment label to every timestamp treats early and late windows as equally informative, even when evidence is unevenly distributed.
Weighting the per-timestamp loss by $\mathrm{IoU}(W^i,S^i)^\beta$ prioritizes windows with higher segment overlap—typically later, more informative windows—while suppressing early low-overlap windows that may share identical inputs yet exhibit label ambiguity during transitions.
This mitigates label interference without introducing look-ahead bias, with larger $\beta$ further concentrating learning on high-overlap windows.

\noindent\textbf{Overall objective.}
The encoder is trained with the sum of both components:
\begin{equation}
\mathcal{L} \;=\; \mathcal{L}^\alpha_{\text{CM}} \;+\; \mathcal{L}_{\text{IoU * CE}}^\beta.
\end{equation}

Notes that textual and audio embeddings $t_i$ and $a_i$ may be aggregated over modality-specific windows $n_t$ and $n_a$ up to $i$ to account for non-synchronous rates.

\subsection{Selective Routing to the VLM Expert}

%\textcolor{red}{Calling the VLM at every timestamp would approximate an upper bound on accuracy, but it is prohibitively expensive in terms of latency and compute for high-cadence streams. At the other extreme, never calling the VLM delegates all decisions to a lightweight encoder that is trained on limited data, has only a bounded past-only window, and lacks the broader social and world knowledge captured by large models.  Moreover, many timestamps are either easy (clear evidence) or genuinely undecidable under the current context; escalating uniformly wastes resources without improving decisions. An adaptive routing strategy is therefore central: it should allocate the VLM only where the expected utility of an expert re-analysis is high, and otherwise let the encoder handle the common case.}
Never invoking the VLM delegates all decisions to a lightweight encoder trained on limited data, lacking the broader social and world knowledge of large models, which degrades performance. Conversely, calling the VLM at every timestamp approximates an accuracy upper bound but is prohibitively costly in latency and compute for high-cadence streams. Moreover, many timestamps are either easy (clear evidence) or genuinely undecidable in the current context, so uniform escalation wastes resources without improving decisions.
Given the encoder’s per-timestamp prediction on the past-only window, routing decides whether to emit the encoder label directly or invoke a VLM expert to reprocess the raw inputs up to time $i$. Upon invocation, the VLM takes the current frame, text from $i!-!N$ to $i$, and a query prompt to produce a refined prediction at time $i$; both the encoder and VLM are trained or fine-tuned.

We consider two routing strategies that use the encoder’s predicted class probability $p_i \in [0.5, 1.0]$: 

\textbf{(1) Model-based adaptive strategy:} a small multilayer perceptron (MLP) takes the encoder’s past probability scores $p^i = [p_{i-N}, \dots, p_i]$ and predicts whether should invoke the VLM to correct the encoder error at time $i$. 

\textbf{(2) Threshold-based rule:} the VLM is invoked when either a label change is detected or the encoder probability falls below a dynamic threshold $\theta^{\text{Enc}}$ that increases with the distance $d=i-t_{\text{vlm}}$ from the last successful VLM-invoked prediction time $t_{\text{vlm}}$:
\begin{equation}
\theta^{\text{Enc}}(d) = 0.5 + \frac{d}{MaxEnc + 1} \cdot 0.5, \quad 1 \leq d \leq MaxEnc + 1.
\end{equation}

Here, $\text{MaxEnc}$ denotes the Maximum Encoder Allowance Step. As $d$ increases, lightweight encoder predictions become more unreliable; if $d=\text{MaxEnc}+1$ we force a VLM call. Setting $\text{MaxEnc}=0$ fixes $\theta^{\text{Enc}}$ at $1$, so the VLM is always invoked; larger $\text{MaxEnc}$ lowers the VLM invocation rate.

\subsection{Deferral Policy}

Some timestamps are genuinely undecidable under the current past-only context. Even after a VLM invocation, the system may defer rather than guess. Similarly, we instantiate two deferral strategies using the VLM model’s probability score $p_i \in [0.5, 1.0]$.
 
\textbf{(1) Model-based adaptive strategy: }A second MLP takes the VLM’s past probability scores $p^i$ and predicts whether the current VLM prediction is correct at time $i$; if not, the system outputs \textit{Defer}. 

\textbf{(2) Threshold-based rule:}
\textit{Defer} is triggered when the VLM confidence falls below a dynamic threshold $\theta^{\text{VLM}}$ that decreases with the distance from the last successful prediction, $d = i - \max(t_{\text{enc}}, t_{\text{vlm}})$.

\begin{equation}
\theta^{\text{VLM}}(d) = 1.0 - \frac{d}{MaxDefer + 1} \cdot 0.5, \quad 1 \leq d \leq MaxDefer + 1.
\end{equation}

Here, $\text{MaxDefer}$ denotes the Maximum Deferral Allowance Steps. As $d$ grows, prolonged deferral becomes harder; if $d=\text{MaxDefer}+1$ we require a prediction (no further deferral). When $\text{MaxDefer}=0$, $\theta^{\text{VLM}}$ is fixed at $0.5$, so deferral is disabled; increasing $\text{MaxDefer}$ allows more deferrals under low-confidence contexts.

%% file: main/experiments.tex
\begin{table*}[t]
\centering
\small
  \caption{Model performance on \textit{streaming social task detection}. Metrics: VLM-Suc = VLM Success Rate, VLM-Defer = VLM Defer Rate, Acc = Accuracy, M-F1 = Macro-F1. The highest values are highlighted; the second-highest values are underlined. VLM-Suc and VLM-Defer are aggregated across datasets for brevity.}  
\begin{tabular}{lllll|cc|cc|cc} \\ 
\hline
&&&&
& \multicolumn{2}{c}{\textbf{MOSI}} 
& \multicolumn{2}{c}{\textbf{MOSEI}} 
& \multicolumn{2}{c}{\textbf{HateClipSeg}}  \\
\cline{6-11}
\textbf{Model}&
\textbf{Latency $\downarrow$}
& \textbf{GPU $\downarrow$}
&
\textbf{VLM-Suc  $\downarrow$} & 
\textbf{VLM-Defer $\downarrow$}  
&\textbf{Acc  $\uparrow$} & \textbf{M-F1  $\uparrow$} &  \textbf{Acc  $\uparrow$} & \textbf{M-F1  $\uparrow$} &  \textbf{Acc  $\uparrow$} & \textbf{M-F1  $\uparrow$} \\
\hline
\hline
OadTR & \SI{0.1}{\second} & 2 GB & 0\% & 0\%  &  71.10 & 68.73 &  73.86 &
64.10 &  63.04 & 62.48 \\ 
LSTR & \SI{0.1}{\second}  & 2 GB  & 0\% & 0\% &  65.56 & 64.55 &  68.97 & 60.82 & 63.21 & 62.75 \\ 
Stream Encoder (ours) & \SI{0.1}{\second} & 2 GB  & 0\% & 0\%  &  71.72 & 70.79 &  76.52 & 65.70 &  64.01 & 63.94 \\

Qwen2.5 &\SI{0.7}{\second}  &  20 GB & 100\% & 0\%   & 79.80 & 77.71 & 84.44 & 72.35 & 67.82 & 67.82 \\

LLaVA-Next &\SI{1.1}{\second}  &  21 GB & 100\% & 0\%  & 80.10 & 78.42 & 83.77& 71.87  & \underline{68.52} & 67.95 \\

Llama-3.2  &\SI{0.8}{\second}  & 27 GB & 100\% & 0\%  &  \underline{80.85} & \underline{79.05} &  \underline{85.35}
& \underline{73.43} &  68.51 & \underline{68.51} \\

StreamSense (Llama-3.2) & \SI{0.3}{\second} & 29 GB & $10 \pm 3\%$ & $17 \pm 3\%$  & \textbf{81.24} & \textbf{79.26} & \textbf{85.64} & \textbf{74.23} &  \textbf{72.10} & \textbf{72.06}  \\
\hline
  \end{tabular}
 \label{tab:streamingsocialtask}
\end{table*}

\begin{table}[t]
\centering
\small
  \caption{Encoder performance under varying loss configurations with $\alpha$ (cross-modal contrastive loss) and $\beta$ (IoU-based cross-entropy loss). Metrics: Acc = Accuracy, M-F1 = Macro-F1. The highest values are highlighted.}  
\begin{tabular}{l|cc|cc|cc} \\ 
\hline
& \multicolumn{2}{c}{\textbf{MOSI}} 
& \multicolumn{2}{c}{\textbf{MOSEI}} 
& \multicolumn{2}{c}{\textbf{HateClipSeg}}  \\
\cline{2-7}

 \textbf{$\alpha$} &\textbf{Acc $\uparrow$} & \textbf{M-F1 $\uparrow$} &  \textbf{Acc $\uparrow$} & \textbf{M-F1 $\uparrow$} &  \textbf{Acc $\uparrow$} & \textbf{M-F1 $\uparrow$} \\
\hline
\hline
  0.00 &    70.07 & 69.31 & 75.42 & 65.47 & 62.79 & 62.74 \\ 
  0.20 &   70.87 & 69.97 & 76.79 & 65.61 & 63.74 & 63.68 \\ 
  0.25 &  \textbf{71.72} & \textbf{70.79} &  \textbf{76.52} & \textbf{65.70} &  \textbf{64.01} & \textbf{63.94} \\ 
0.30 &   71.21  & 70.25  & 76.45  & 65.47  & 63.93  & 63.90 \\ 
\hline
\hline
  \textbf{$\beta$} &\textbf{Acc $\uparrow$} & \textbf{M-F1 $\uparrow$} &  \textbf{Acc $\uparrow$} & \textbf{M-F1 $\uparrow$} &  \textbf{Acc $\uparrow$} & \textbf{M-F1 $\uparrow$} \\
\hline
\hline
 0.0 &   70.92 & 69.84 &  73.76 & 63.92 & 62.72 & 62.71 \\ 
 0.8 &   70.92 & 69.84 & 76.02  & 65.61 & 63.48  & 63.38 \\ 
 1.0 &  \textbf{71.72} & \textbf{70.79} &  \textbf{76.52} & \textbf{65.70} &  \textbf{64.01} & \textbf{63.94} \\ 
  1.2 &  70.70 & 69.86 & 76.32 & 65.65  & 63.91  & 63.89 \\ 
\hline
  \end{tabular}
 \label{tab:transformr_loss}
\end{table}

\begin{table*}[t]
\centering
\small
  \caption{StreamSense performance under varying VLM-invocation and prediction deferral strategies. Metrics: VLM-Suc = VLM Success Rate, VLM-Defer = VLM Defer Rate, Acc = Accuracy, M-F1 = Macro-F1. The highest values are highlighted. VLM-Suc and VLM-Defer are aggregated across datasets for brevity.}  
\begin{tabular}{lllll|cc|cc|cc} \\ 
\hline
&&&&
& \multicolumn{2}{c}{\textbf{MOSI}} 
& \multicolumn{2}{c}{\textbf{MOSEI}} 
& \multicolumn{2}{c}{\textbf{HateClipSeg}}  \\
\cline{6-11}
 \textbf{Strategy} &  \textbf{Latency $ \downarrow $} &  \textbf{GPU $ \downarrow $} & 
\textbf{VLM-Suc $ \downarrow $} &  \textbf{VLM-Defer $ \downarrow $}  
&\textbf{Acc $\uparrow$} & \textbf{M-F1 $\uparrow$} &  \textbf{Acc $\uparrow$} & \textbf{M-F1 $\uparrow$} &  \textbf{Acc $\uparrow$} & \textbf{M-F1 $\uparrow$} \\

\hline
\hline

Model &  \SI{0.4}{\second} & 29 GB & $23 \pm 1\%$ & $10 \pm 5\%$ & 76.47 & 75.26 &  84.32 & 72.56 &  68.57 & 68.49  \\

% Threshold (\textit{No VLM}) & 0.40  & 0\% & $30 \pm 16\%$  & 73.13 & 67.97 &78.65 & 65.10 & 64.02 & 63.94 \\

% Threshold (\textit{No Defer}) & 0.26 & $20 \pm 4\%$ & 0\%   & 78.46 & 77.00 &83.13 & 73.16 & 68.70 &68.67  \\

Threshold (\textit{No VLM}) & \SI{0.5}{\second} & 29 GB & 0\% & $43 \pm 10\%$ & 77.45 &77.29 &84.48 & 67.86 &  67.70 & 67.57 \\

Threshold (\textit{No Defer}) & \SI{0.3}{\second} & 29 GB & $20 \pm 4\%$ & 0\%   & 78.35 & 76.89 & 82.98 & 72.93 & 68.70 &68.67   \\

Threshold (\textit{Allow Both}) & \SI{0.3}{\second} & 29 GB & $10 \pm 3\%$ & $17 \pm 3\%$  & \textbf{81.24} & \textbf{79.26} & \textbf{85.64} & \textbf{74.23} &  \textbf{72.10} & \textbf{72.06}    \\

\hline
  \end{tabular}
 \label{tab:VLM_invocation_delay}
\end{table*}

\begin{table*}[ht]
\centering
\small
\caption{StreamSense predictions on MOSI dataset under different settings of the threshold-based strategy; only text displayed.}
\begin{tabular}{p{8.7cm}|c|c|c|c}
\hline
\textbf{Text} & \textbf{Ground Truth} & \textbf{\textit{No VLM}} & \textbf{\textit{No Defer}} & \textbf{\textit{Allow Both}} \\ \hline \hline
The action is really, really well directed. It's very reminiscent of District 9.  & \textit{positive} & \textit{positive} & \textit{positive} & \textit{positive} \\ \hline
It's very reminiscent of District 9. It's like someone watched & \textit{negative} & \textit{positive} & \textit{positive} & \textit{defer} \\ \hline
 It's like someone watched the third act of District 9 a bunch of times, and like& \textit{negative} & \textit{positive} & \textit{negative} & \textit{negative} \\ \hline
\end{tabular}

\label{table:demo_example}
\end{table*}

\begin{table*}[t]
\centering
\small
  \caption{StreamSense performance under varying modality. Metrics: VLM-Suc = VLM Success Rate, VLM-Defer = VLM Defer Rate, Acc = Accuracy, M-F1 = Macro-F1. Top values highlighted. VLM-Suc and Defer aggregated across datasets for brevity.}  
\begin{tabular}{lllll|cc|cc|cc} \\ 
\hline
&&&&
&  \multicolumn{2}{c}{\textbf{MOSI}} 
& \multicolumn{2}{c}{\textbf{MOSEI}} 
& \multicolumn{2}{c}{\textbf{HateClipSeg}}  \\
\cline{6-11}
\textbf{Modality} &  \textbf{Latency $ \downarrow $} &   \textbf{GPU $ \downarrow $} &
\textbf{VLM-Suc $ \downarrow $} & 
\textbf{VLM-Defer $ \downarrow $} 
&\textbf{Acc $ \uparrow $} & \textbf{M-F1 $ \uparrow $} &  \textbf{Acc $ \uparrow $} & \textbf{M-F1 $ \uparrow $} &  \textbf{Acc $ \uparrow $} & \textbf{M-F1 $ \uparrow $} \\
\hline
\hline
V & \SI{0.4}{\second} & 28 GB & $12 \pm 4\%$ & $24 \pm 8\%$  &   68.64 & 40.70 &  80.93 & 45.51 &  66.19 & 66.10 \\

T & \SI{0.3}{\second} & 28 GB & $10 \pm 3\%$ & $17 \pm 5\%$ &  78.07 & 74.02 &  82.92 & 73.58 &  64.28 & 64.27   \\

V, T, A & \SI{0.3}{\second} & 29 GB & $10 \pm 3\%$ & $17 \pm 3\%$  & \textbf{81.24} & \textbf{79.26} & \textbf{85.64} & \textbf{74.23} &  \textbf{72.10} & \textbf{72.06}  \\
\hline
  \end{tabular}
 \label{tab:modality}
\end{table*}

\begin{figure}[t]
  \centering
  \begin{minipage}{0.41\textwidth}
    \centering
    \includegraphics[width=\linewidth]{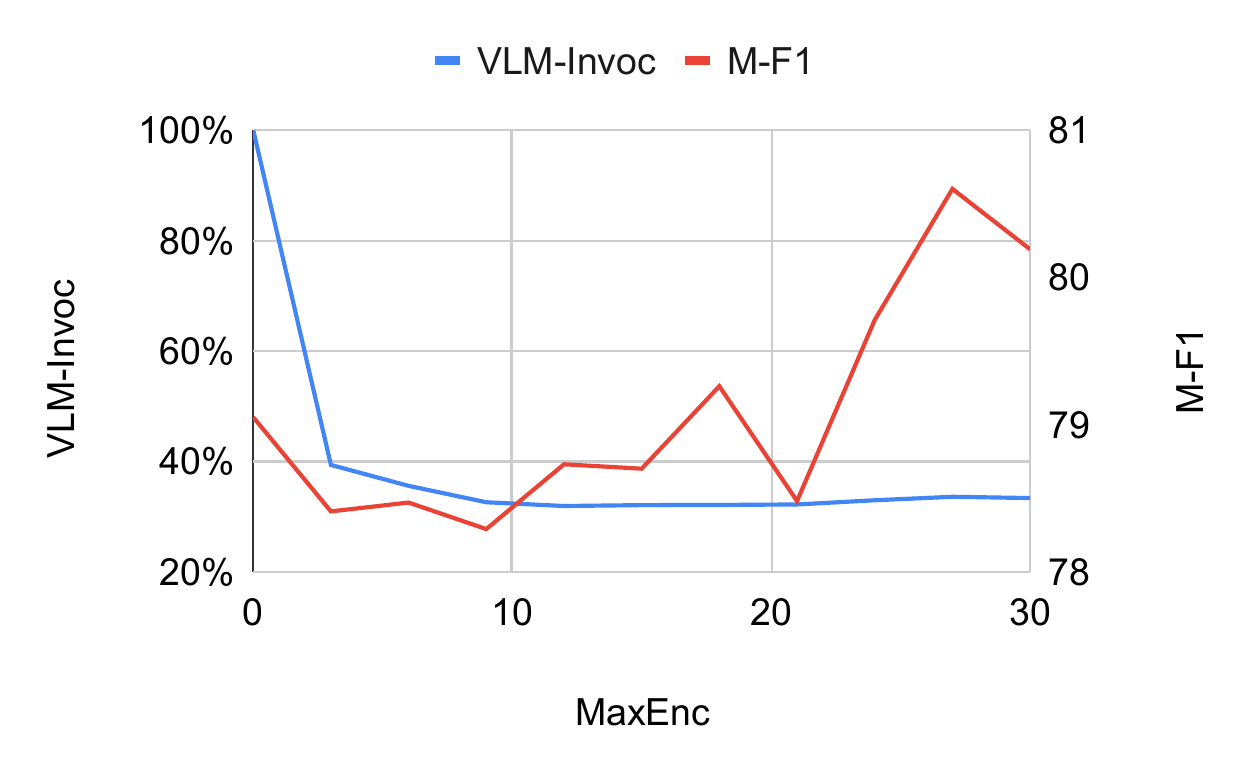}
  \end{minipage}
  \hfill
  \begin{minipage}{0.41\textwidth}
    \centering
    \includegraphics[width=\linewidth]{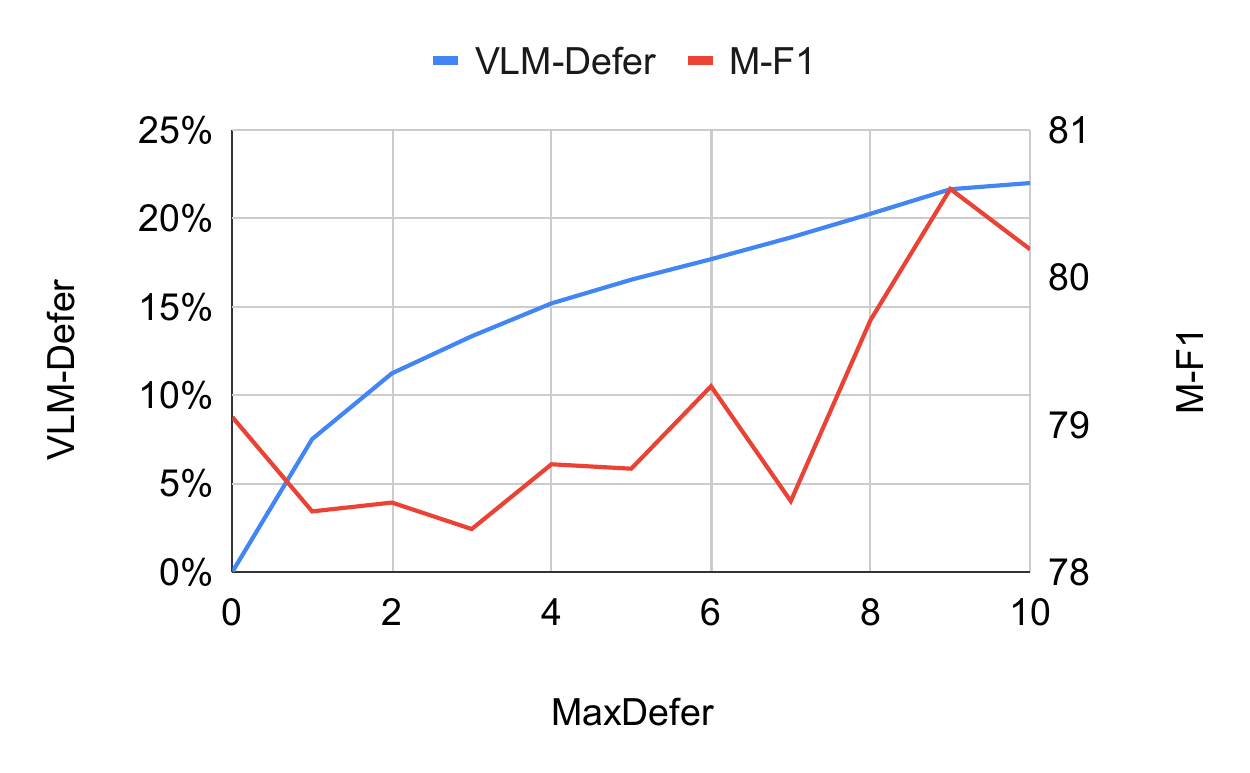}
  \end{minipage}
    
    \caption{StreamSense model performance for MOSI dataset under the threshold-based VLM-Invocation and Prediction Defer strategy, evaluated across varying MaxEnc (0–30) and MaxDefer (0–10). MaxEnc: Maximum Encoder Allowance Steps, MaxDefer: Maximum Deferral Allowance Steps . Metrics: VLM-Invoc = VLM Invocation Rate, VLM-Defer = VLM Defer Rate, M-F1 = Macro-F1.}
    \label{fig:MaxDefer_MaxEnc}
\end{figure}

% \begin{figure}[t]
%   \centering
 
%   \begin{minipage}{0.48\textwidth}
%     \centering
%     \includegraphics[width=\linewidth]{images/ElapsedTime.pdf}
%   \end{minipage}
   
%     \caption{VLM Defer Rate (VLM-Defer) across varying segment-level elapsed time (SegElapsedTime) for StreamSense model on the MOSI dataset, comparing MaxEnc = 12 and MaxDefer = 5 (optimal).}
%      \label{fig:elapsed_time}
% \end{figure}

\subsection{Experimental Settings}

\subsubsection{Datasets.} We train and evaluate \textsf{StreamSense} on three streaming social task datasets.

\textbf{MOSI} (CMU-MOSI \cite{zadeh2016mosi}) is a multimodal sentiment intensity corpus with 2,199 segments from 93 YouTube monologue videos. Each segment is annotated with a sentiment score in [$-3$, $+3$]. We binarize sentiment as \textit{negative} ($y_i=0$) for scores $<0$ and \textit{positive} ($y_i=1$) for scores $\ge 0$, excluding unannotated parts of the videos.

\textbf{MOSEI} (CMU-MOSEI \cite{zadeh2018cmumosei}) comprises 23,453 annotated segments from 3,228 YouTube videos. Each segment is labeled with a sentiment score in [$-3$, $+3$] and six emotion categories. Following MOSI, we treat scores $<0$ as \textit{negative} ($y_i=0$) and scores $\ge 0$ as \textit{positive} ($y_i=1$), excluding unannotated portions. Due to the public release not including all raw videos and some links being broken, our final MOSEI subset comprises 12,803 segments from 2,189 downloaded videos.

HateClipSeg \cite{wang2025hateclipseg} is a segment-level hate video dataset with researcher-defined boundaries that ensure consistent and complete opinions. It contains 11,714 segments from 435 YouTube and Bitchute videos labeled as \textit{Normal} or one of five Offensive categories (\textit{Hateful}, \textit{Insulting}, \textit{Sexual}, \textit{Violence}, \textit{Self-Harm}). We map \textit{Normal} to $y_i=0$ and all offensive categories to $y_i=1$.
%While there are other hateful video datasets, we have excluded them in our study because segment boundaries are subjectively determined by annotators, making segment-level annotations unreliable for our setting.

\subsubsection{Preprocessing.}
Data are split at the video level into 80\% training and 20\% testing. The features are sampled every second, and the inference interval is also set to $s=1$\,s. The context window size is $N=32$, covering the most recent 32\,s of content. Unimodal features are extracted using pretrained models: ViT-Large \cite{dosovitskiy2020image} for vision, BERT \cite{devlin2019bert} for text, and Wav2Vec2 \cite{baevski2020wav2vec} for audio. Based on encoder performance on MOSI, the optimal unimodal windows are $n_t=2$ s for text and $n_a=4$ s for audio, with vision excluded.

\subsubsection{Baselines.} We evaluate six baselines. For Encoder-only, we evaluate our Stream Encoder and two visual online action recognition models, OadTR~\cite{wang2021oadtr} and LSTR~\cite{xu2021long}, adapted to multi-modal inputs by concatenating the three uni-modal features. For VLM-only, we consider Qwen2.5-VL-7B-Instruct (Qwen2.5)~\cite{qwen2.5-VL}, LLaVA-NeXT-Video-7B (LLaVA-Next)~\cite{zhang2024llava}, and Llama-3.2-11B-Vision (Llama-3.2) ~\cite{dubey2024llama}. These baselines highlight the effect of selective routing.

\subsubsection{Implementation Details.} All experiments are conducted on a single NVIDIA RTX A6000 GPU with FP16 precision. Our Stream Encoder employs three uni-modal 2-layer encoders ($l_{enc}=2$), a 2-layer fusion encoder ($l_{fus\_enc}=2$), and a 2-layer MLP. The VLM is a pretrained Llama-3.2-11B-Vision (LLaMA-3.2) \cite{dubey2024llama}. Both the Encoder and the VLM are trained on streaming social detection data for 10 epochs; the Encoder uses $s=1$\,s sampling, and the VLM uses $s=4$\,s sampling to reduce training time. The best epoch is selected for selective routing. For VLM invocation and prediction deferral, the model-based strategy uses a 2-layer MLP; the threshold-based strategy sets $MaxEnc=18$ and $MaxDefer=6$ across datasets. Unless stated otherwise, we use the threshold-based method.

\subsubsection{Evaluation Metrics.}
The SSTD tasks are evaluated at fixed intervals $s=1$\,s using Accuracy (Acc) and Macro-F1 (M-F1), with M-F1 as the primary metric. Efficiency is measured by VLM-Suc (successful VLM invocations) and VLM-Defer (deferred invocations), whose sum defines VLM-Invoc (overall VLM usage). We report latency (seconds per prediction, counting each deferral as a 1-second delay) and GPU usage to reflect deployment cost.

\subsection{Main Results}

Table~\ref{tab:streamingsocialtask} summarizes the SSTD performance across three datasets and benchmark models. Among lightweight models, our stream encoder had achieved the best performance across all datasets. The stream encoder adopts a multimodal design (separate branches per modality with cross-modal contrastive alignment and multi-modal fusion) and uses an IoU-based classification loss that targets label interference at segment boundaries. In contrast, OadTR and LSTR—designed for visual-only or synchronous online action recognition—perform poorly when naively extended to multimodality, indicating that simple concatenation is inadequate for social tasks where key cues are textual or acoustic.

We observe that all VLMs consistently outperform lightweight models across datasets but incur substantially higher latency, underscoring the accuracy–latency trade-off of large models in streaming social understanding. This observation suggests that a pure VLM represents an accuracy upper bound among single-component systems but is impractical for low-latency or cost-sensitive deployment; among evaluated models, Llama-3.2 performed best and was selected as the base VLM.

\textsf{StreamSense}, which combines the lightweight encoder with threshold based VLM invocation and prediction deferral, achieves the best overall trade-off: it attains the highest performance while invoking the VLM on only 27\% of timestamps, with 17\% corresponding to deferrals. It yields a moderate latency of \SI{0.3}{\second}. By explicitly deferring when context is insufficient, \textsf{StreamSense} surpasses the VLM-only system despite much lower VLM usage. This effect is most pronounced on HateClipSeg, where deferral is especially beneficial (M-F1 72.06 for StreamSense versus 68.51 for VLM-only), consistent with the notion that hate speech requires a broader context than sentiment.This result shows that selective routing with deferral is essential in streaming settings, improving accuracy by avoiding premature decisions while reducing cost and latency via judicious VLM use.

\subsection{Ablation Study on the Encoder Loss}
We ablate the encoder’s two loss components: cross-modal contrastive (weight $\alpha$) and IoU-based cross-entropy (weight $\beta$), to assess their contributions with varying hyperparameters (Table~\ref{tab:transformr_loss}).

For the cross-modal contrastive term, incorporating it ($\alpha>0$) consistently improves M-F1 on MOSI and HateClipSeg compared to removing it ($\alpha=0$). On MOSEI, the gains are limited, consistent with the task being largely text-dominated (Table~\ref{tab:modality} shows multimodal exceeds text-only by only 0.32 M-F1). The best setting is $\alpha=0.25$. From the ablation results, we noted that (i) alignment helps when non-text cues contribute meaningfully (MOSI, HateClipSeg), but offers little headroom when text already carries most signal (MOSEI); (ii) using a modest $\alpha$ avoids over-constraining representations while retaining the robustness benefits of alignment, so we keep $\alpha{=}0.25$ in subsequent experiments.

For the IoU-based cross-entropy, using it ($\beta>0$) significantly outperforms standard cross-entropy ($\beta=0$) on all datasets, with an average M-F1 increase of 1.32. This supports the hypothesis that label interference around segment boundaries is a consistent challenge in streaming social tasks and that weighting by temporal overlap mitigates it. The best setting is $\beta=1$. Interesting, we note that (i) boundary-aware supervision is necessary for streaming evaluation derived from segment labels and should be treated as a default training objective in this setting; (ii) the consistent gains across datasets justify fixing $\beta{=}1$ for the main results.

%\textit{Takeaway.} Cross-modal alignment is situationally beneficial and should be tuned to the modality balance of the dataset, whereas IoU-weighted classification yields uniform improvements and is essential for stabilizing training under segment-level supervision.

%\subsection{Ablation Study for Encoder Loss}

%We perform an ablation study on the Encoder loss coefficients (Table~\ref{tab:transformr_loss}) to assess our design and determine the optimal values.

%The encoder classifier uses two losses: a cross-modal contrastive loss controlled by $\alpha$ and an IoU-based cross-entropy loss controlled by $\beta$.
%Results show that incorporating cross-modal contrastive loss ($\alpha>0$) consistently improves performance on MOSI and HateClipSeg compared to its absence ($\alpha=0$). Gains on MOSEI are limited, likely because the text modality already dominates performance, and aligning other modalities contributes minimally, as shown in Table~\ref{tab:modality}, where multimodal M-F1 improves by only 0.32 over text-only. The optimal $\alpha$ is 0.25.
%Results indicate that IoU-based cross-entropy loss ($\beta>0$) significantly improves performance over standard cross-entropy ($\beta=0$) across all datasets, with an average M-F1 gain of 1.32. This demonstrates that label interference is a consistent issue across social video datasets and can be mitigated using our IoU-based classification loss. The optimal value is $\beta=1$.

\subsection{Ablation Study on VLM Invocation and the Prediction Deferral Strategy}
Table~\ref{tab:VLM_invocation_delay} compares the VLM invocation and deferral strategies.

\subsubsection{Model-based vs. threshold-based.}
The model-based method (33\% VLM usage; \SI{0.4}{\second} latency) outperforms the Encoder-only setting but remains inferior to the VLM-only system. This suggests that relying solely on the model’s probability scores, without human insights, is insufficient for reliable escalation and deferral. In contrast, our deliberated threshold-based method provides stronger control over when to invoke the VLM and when to defer.

\subsubsection{Threshold-based settings.}
We evaluate three configurations: \textit{No VLM} (the Encoder’s probabilities drive deferral; VLM-Suc $=0\%$), \textit{No Defer} (VLM-Defer $=0\%$), and \textit{Allow Both}. Using the Encoder’s probabilities to decide deferral performs worse than not deferring at all, indicating that these scores should not be used to trigger deferral. Allowing both mechanisms yields the best results: invoking the VLM on only 20\% of timestamps without deferral and at a latency of \SI{0.3}{\second} already approaches the performance of the VLM-only system, and enabling deferral further improves results, surpassing the VLM-only system while keeping VLM usage low.

\subsubsection{Effect of $MaxEnc$ and $MaxDefer$.}
In the threshold-based \textit{Allow Both} setting (Figure~\ref{fig:MaxDefer_MaxEnc}), we vary Maximum Encoder Allowance Steps ($MaxEnc$) and Maximum Defer Allowance Steps ($MaxDefer$). Because $MaxEnc$ primarily controls VLM-Invoc and $MaxDefer$ controls VLM-Defer, we report M-F1 with the corresponding rate on each sweep. A $3:1$ ratio of $MaxEnc:MaxDefer$ performs best; fixing this ratio and sweeping $MaxEnc$ from 0–30 ($MaxDefer$ 0–10) shows overall M-F1 remains stable between 78.5 and 80.5, indicating robustness. Increasing $MaxEnc$ reduces VLM-Invoc by permitting more Encoder emissions; increasing $MaxDefer$ raises VLM-Defer. Within $MaxEnc =$ 12 - 24 and $MaxDefer =$ 4 – 8, both invocation and deferral remain at acceptable levels and are consistent across datasets. Accordingly, we set $MaxEnc = 18$ and $MaxDefer = 6$ in all reported experiments.

\subsubsection{Qualitative illustration.}
Table~\ref{table:demo_example} presents three consecutive timestamps from one video. The first and last have sufficient context; the last is sarcastically negative. \textit{No VLM} fails on the last timestamp, showing the benefit of VLM invocation on ambiguous cases. The middle timestamp is under-informative; both \textit{No VLM} and \textit{No Defer} misclassify it as positive. Although \textit{No VLM} permits deferral, the Encoder’s probabilities are unreliable for this decision. Only \textit{Allow Both} correctly identifies insufficient context and defers until more evidence becomes available.

\subsection{Ablation Study on Modality}
We report unimodal and multimodal variants of \textsf{StreamSense} in Table~\ref{tab:modality}. Audio-only results are omitted because audio is only supported by the Encoder. Multimodal \textsf{StreamSense} consistently outperforms unimodal variants, with the largest gains on MOSI and HateClipSeg, where the combined visual, textual, and audio (V, T, A) model clearly exceeds the best unimodal setting. On MOSEI, improvements are minor, consistent with text alone providing a sufficient signal. Among unimodal models, text is the most informative modality for sentiment. In contrast, visual cues are more critical for hate speech detection, likely because HateClipSeg does not require strict transcripts, and many segments lack text. Overall, the multimodal model remains robust, even when some modalities are incomplete, by effectively leveraging the available inputs.

%\subsection{Ablation Study for Modality}
%We present  multimodal and unimodal results of StreamSense in Figure~\ref{tab:modality} (audio excluded, as only Encoder supports it) to illustrate the effect of modality on performance.

%Multimodal StreamSense consistently outperform unimodal ones, with the effect most pronounced on MOSI and HateClipSeg, where the combined visual, textual, and audio (V, T, A) model markedly exceeds the best unimodal performance. For MOSEI, gains are smaller, as text alone provides sufficient information. Among unimodal models, text is most informative for sentiment tasks, while visual cues are more important for hate speech detection, likely because HateClipSeg lacks strict transcript requirements, leading to missing text in many segments. Overall, our multimodal model remains robust on datasets even with incomplete modalities by effectively leveraging information from the available modalities. 

%% file: main/conclusion.tex
We introduce streaming social task detection, formalize its setting, and propose \textsf{StreamSense}, a selective-routing framework that couples a lightweight, modality-flexible encoder with a VLM. The encoder is trained via cross-modal contrastive alignment and an IoU-weighted classification loss to reduce label interference, while the VLM is selectively invoked or deferred under insufficient context. This design surpasses VLM-only baselines with lower VLM usage, remains model-agnostic, and supports practical streaming social media monitoring. A limitation is its reliance on past-only evidence, which may bias predictions under limited global context; in practice, a short decision buffer is advisable.

%In conclusion, we present one \textit{streaming social task detection} model, defining the task and its unique challenges. To achieve high detection performance with low inference cost, we propose a selective-routing streaming detector (StreamSense) that combines the efficiency and modality flexibility of a lightweight stream encoder with the strong social task performance of a vision-language model (VLM). The Encoder is trained with cross-modal contrastive loss for modality alignment and IoU-based classification loss to mitigate label interference. The VLM is selectively invoked for difficult cases and can defer predictions with insufficient context. Our approach outperforms fully VLM-based methods with substantially lower VLM usage and is model-agnostic, allowing integration with more advanced VLMs in the future.
%It offers a strong foundation for \textit{streaming social task detection} with wide applications in live social media monitoring. However, since it relies only on historical content, it may exhibit bias due to limited global context. Therefore, real-world applications should include a buffer period before making final decisions.